\documentclass{article}

\usepackage{microtype}
\usepackage{graphicx}
\usepackage{subfigure}
\usepackage{booktabs} % for professional tables

\usepackage{multirow}

\usepackage{hyperref}

\usepackage[accepted]{icml2023}

\usepackage{amsmath}
\usepackage{amssymb}
\usepackage{mathtools}
\usepackage{amsthm}

\usepackage[capitalize,noabbrev]{cleveref}

\theoremstyle{plain}

\theoremstyle{definition}

\theoremstyle{remark}

\usepackage[textsize=tiny]{todonotes}

\icmltitlerunning{MOLE: MOdular Learning FramEwork via Mutual Information Maximization}

\begin{document}

\twocolumn[
\icmltitle{MOLE: MOdular Learning FramEwork via Mutual Information Maximization}

% It is OKAY to include author information, even for blind
% submissions: the style file will automatically remove it for you
% unless you've provided the [accepted] option to the icml2023
% package.

% List of affiliations: The first argument should be a (short)
% identifier you will use later to specify author affiliations
% Academic affiliations should list Department, University, City, Region, Country
% Industry affiliations should list Company, City, Region, Country

% You can specify symbols, otherwise they are numbered in order.
% Ideally, you should not use this facility. Affiliations will be numbered
% in order of appearance and this is the preferred way.
\icmlsetsymbol{equal}{*}

\begin{icmlauthorlist}
\icmlauthor{Tianchao Li}{equal,yyy}
\icmlauthor{Yulong Pei}{equal,yyy}
%\icmlauthor{Firstname3 Lastname3}{comp}
%\icmlauthor{Firstname4 Lastname4}{sch}
%\icmlauthor{Firstname5 Lastname5}{yyy}
%\icmlauthor{Firstname6 Lastname6}{sch,yyy,comp}
%\icmlauthor{Firstname7 Lastname7}{comp}
%\icmlauthor{}{sch}
%\icmlauthor{Firstname8 Lastname8}{sch}
%\icmlauthor{Firstname8 Lastname8}{yyy,comp}
%\icmlauthor{}{sch}
%\icmlauthor{}{sch}
\end{icmlauthorlist}

\icmlaffiliation{yyy}{Department of Mathematics and Computer Science,  Eindhoven University of
Technology,  Eindhoven,  the Netherland}
%\icmlaffiliation{comp}{Company Name, Location, Country}
%\icmlaffiliation{sch}{School of ZZZ, Institute of WWW, Location, Country}

\icmlcorrespondingauthor{Tianchao Li}{litianchao1996@outlook.com}
\icmlcorrespondingauthor{Yulong Pei}{y.pei.1@tue.nl}

% You may provide any keywords that you
% find helpful for describing your paper; these are used to populate
% the "keywords" metadata in the PDF but will not be shown in the document
\icmlkeywords{Machine Learning, ICML}

\vskip 0.3in
]

% this must go after the closing bracket ] following \twocolumn[ ...

% This command actually creates the footnote in the first column
% listing the affiliations and the copyright notice.
% The command takes one argument, which is text to display at the start of the footnote.
% The \icmlEqualContribution command is standard text for equal contribution.
% Remove it (just {}) if you do not need this facility.

%\printAffiliationsAndNotice{}  % leave blank if no need to mention equal contribution
\printAffiliationsAndNotice{\icmlEqualContribution} % otherwise use the standard text.

\begin{abstract}
%This document provides a basic paper template and submission guidelines. Abstracts must be a single paragraph, ideally between 4--6 sentences long. Gross violations will trigger corrections at the camera-ready phase.

This paper is to introduce an asynchronous and local learning framework for neural networks, named Modular Learning Framework (MOLE). This framework modularizes neural networks by layers, defines the training objective via mutual information for each module, and sequentially trains each module by mutual information maximization. MOLE makes the training become local optimization with gradient-isolated across modules, and this scheme is more biologically plausible than BP. We run experiments on vector-, grid- and graph-type data. In particular, this framework is capable of solving both graph- and node-level tasks for graph-type data. Therefore, MOLE has been experimentally proven to be universally applicable to different types of data.
%Therefore, MOLE is experimentally proven to be generally applicable to diverse types of data.
\end{abstract}

\section{Introduction}
\label{introduction}
Deep Neural Networks (DNNs) have attracted attention in many domains over the last decade as a result of their empirical success in a variety of applications, such as image classification \cite{sornam2017survey}, natural language processing \cite{otter2020survey}, speech recognition \cite{nassif2019speech} and anomaly detection \cite{pang2021deep}. Backpropagation \citep[BP,][]{rumelhart1986learning}, as the dominant and mature algorithm of training deep neural networks, plays a central role in the success of DNNs. BP is an end-to-end or global optimization algorithm where all the weight parameters are automatically updated by backpropagating the error gradient from the loss function to each layer until the input layer. BP excessively depends on labeled data. Specifically, each input sample for DNNs requires a correct label to compute the loss via the loss function and DNNs cannot feed unlabeled samples. To ensure good generalization of DNNs, BP requires large labeled datasets. In contrast, humans learn from a few shots of labeled samples to recognize the new category: they summarize the special features of this category from the given samples and recognize the unobserved samples using summary knowledge. Therefore, BP makes DNNs, designed by imitating human neural networks in the bio-brain, biologically implausible. Furthermore, the bio-brain is very modular and learns mainly based on local information \cite{caporale2008spike}.

BP brings other drawbacks to DNNs. 
%To better solve the complex task, DNNs stack multi-layers with large amounts of parameters. 
During the training, BP needs to compute many gradients to update the corresponding parameters. As all parameters, activations, and gradients of the DNN need to fit into a processing unit's working memory, BP creates a considerable amount of memory when the DNN stacks multi-layers. In addition, BP uses the chain rule to compute the gradients of each layer, namely the parameter's gradient depends on the following parameters in the DNN. So BP leads that implicit correlations across layers would exist, although the parameters only determine the output of their layers. Therefore, the outcome-based internal interpretation and exploration are not reliable enough. Therefore, the gradient-isolated training algorithm or method will bridge this issue.
%For example, after training, exploring the internal output of Convolutional Neural Networks and interpreting how each kernel works.

To relieve these drawbacks from BP, we propose a neural network training framework, named Modular Learning Framework (MOLE). Using this framework, we modularize the neural network by layers with fewer parameters per module. Then we define the training objective for different groups of modules, and each module is sequentially and independently trained by optimizing its objective. Furthermore, we experimentally demonstrate MOLE's universal applicability for various types of data, and we explain the reason for the gap between MOLE and BP. Finally, we compare MOLE with other alternatives to BP and point out potentially feasible improvements.
%{\footnotesize
%\begin{verbatim}
%dvips -Ppdf -tletter -G0 -o paper.ps paper.dvi
%ps2pdf paper.ps
%\end{verbatim}}

\section{Preliminary}
Mutual information (MI) is a measure of the amount of information that one random variable contains about another random variable \cite{cover1991entropy}. A dual interpretation of MI is the reduction in the uncertainty of one random variable due to the knowledge of the other \cite{alajaji2018introduction}. MI has the form for two variables $X$ and $Z$,
\begin{equation}
\small
I(X;Z)= \int_{\mathcal{X}\times\mathcal{Z}} \log \frac{\mathrm{d}\mathbb{P}_{XZ}}{\mathrm{d}\mathbb{P}_X \otimes \mathbb{P}_Z}\mathrm{d}\mathbb{P}_{XZ,} 
\end{equation}
where $\mathbb{P}_{XZ}$ is the joint probability distribution, and $\mathbb{P}_X=\int_{\mathcal{Z}}\mathrm{d}\mathbb{P}_{XZ}$ and $\mathbb{P}_Z=\int_{\mathcal{X}}\mathrm{d}\mathbb{P}_{XZ}$ are the marginal probability distributions. Specifically, MI measures the similarity between two variables' spaces $\mathcal{X}$ and $\mathcal{Z}$ based on the marginal and joint distributions of two variables $\textit{X}$ and $\textit{Z}$. In most settings, these distributions are unknown and would be estimated from the given samples. However, the sample data in deep learning is high-dimensional and even has certain structural properties. Accordingly, the distribution estimation of such data is ill-posed \cite{vapniknature}. Therefore, consistent and precise mutual information estimators on high-dimensional even structural data were proposed by detouring from the distribution estimation.

Some MI estimators were proposed without estimating the underlying data distribution. \citet{giraldo2014measures} proposes a non-parametrical matrix-based MI estimator for vector-type data by defining functionals on normalized positive definite matrices. \citet{yu2021measuring} leverage the above matrix-based estimator and propose a differentiable and statistically more powerful normalized MI estimator. With the success of DNNs, the parametrical and scalable neural estimator is designed for MI estimation with good consistency. MINE \cite{belghazi2018mutual} uses the dual representation of KL-divergence (Donsker-Varadhan representation \cite{donsker1983asymptotic}) to construct the DNN and its expressive power guarantees that the estimation of MINE can be arbitrarily close to the true MI on high-dimensional data. However, MINE does not have a good capability to capture the structural MI. %(e.g. spatial locality in grid-like data and topological connectivity in graph data).
%, defined as:
%\begin{equation}
   % \widehat{I(X;Z)}_n = \sup_{\theta\in\Theta}\mathbb{E}_{\mathbb{P}_{XZ}}^{(n)}[T_{\theta}]-\log(\mathbb{E}_{\mathbb{P}_{X}^{(n)}\otimes \hat{\mathbb{P}}_Z^{(n)}}[e^{T_{\theta}}]),
%\end{equation}
%where the expectations are estimated using empirical samples, and $T_{\theta}$ os tj
DIM \cite{hjelm2018learning} leverages the local MI using the average MI between the high-level vector and local patches of grid-like data and matches representations to a prior distribution by adversarial learning. DGI \cite{velickovic2019deep} extends the ideas of DIM to the graph domain. %: summarize the graph embedding as the high-level vector via a read-out function, and maximize MI between this vector and the representation by discriminating the input graph from the negative samples. 
GMI \cite{peng2020graph} captures the topological connectivity better than DGI because GMI explicitly captures the correlation between input features and hidden vectors of both nodes and edges by defining the topology-aware MI.

One of the extensions of MI is Information Bottleneck (IB). In line with the information-theoretic concept, the goal of DNNs aligns with that of IB, namely finding the trade-off between compression and prediction \citep{tishby2000information}. Formally, given an input $X\in \mathcal{X}$, the internal representation $T\in \mathcal{T}$ and the label $Y\in\mathcal{Y}$, the optimal representation $T$ should be derived from the trade-off the complexity (rate) of the representation, $I(T; X)$, and the amount of preserved relevant information, $I(T; Y)$ \cite{tishby2015deep}. Therefore, IB for DNNs is represented as the maximization of the Lagrangian optimization:
\begin{equation}\label{equ:ib}
    \mathcal{L}_{IB}= I(T;Y)-\beta I(T;X),
\end{equation}
where $\beta$ is the positive Lagrange multiplier. Previous works \cite{shamir2010learning, wang2020infobert, wu2020graph} have proved that IB can improve the robustness of DNNs.

MI derives from the communication system \cite{shannon1948mathematical}. A communication system consists of five components: information source, transmitter, information channel, receiver, and destination. The information source generates messages or a sequence of messages ($X_1$).  The transmitter (also known as the encoder) encodes the message from the information source into a signal ($X_2$) that is suitable for transmission over the information channel. The information channel transmits the signal $X_2$ and outputs the arrival signal $X_3$ to the following receiver. The receiver (also known as the decoder) decodes the information transmitted via the channel as the original information as much as possible, and the output of the decoder is denoted as $X_4$. Assuming all is right with the world, the receiver will have the ability to decode the signal and recover the original message. The data flow from $X_1$ to $X_4$ in the communication system forms a Markov chain: $ X_1\rightarrow X_2 \rightarrow X_3 \rightarrow X_4$.
%\begin{align}
%\small
%    X_1\rightarrow X_2 \rightarrow X_3 \rightarrow X_4.
%\end{align}
\citet{cover1991entropy} defines Data Processing Inequality(DPI) for a Markov chain, so the DPI of communication systems is
%\begin{equation}\label{DPI_cs}
     %H(X_1)\geq I(X_2;X_1)\geq %I(X_3;X_1)\geq I(X_4;X_1),
%\end{equation}
\begin{equation}\label{DPI_cs}
\small
    I(X_2;X_1)\geq I(X_3;X_1)\geq I(X_4;X_1).
\end{equation}
%where $H(X_1)$ represents the Shannon entropy of $X_1$. 
We propose MOLE in Section \ref{methodology} inspired by the mechanism of communication systems because the data flow in communication systems is significantly similar to DNNs'. 
%We are inspired by the mechanism of the communication system and we propose MOLE in Section \ref{methodology}.

\section{Modular Learning Framework}
\label{methodology}
\begin{figure*}[ht]
\centering
\includegraphics[width=0.8\textwidth]{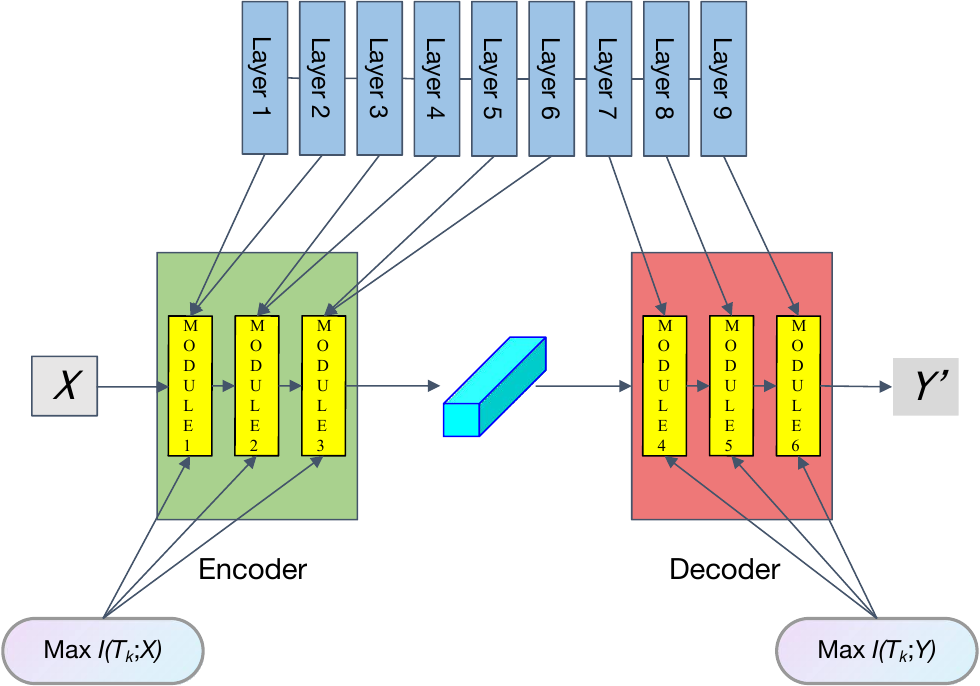}
\caption{The training pipeline of MOLE is shown. $X$ is the input, $Y$ is the label, $Y'$ is the final output, and $T_k$ represents the output of $k$-th module. The top part is a DNN stacking 9 layers, the middle is MOLE, and the bottle is the training objective of each module where $I(*;*)$ represents the mutual information.}
\label{pipline}

\end{figure*}

DNNs process data in a manner quite similar to communication systems. In DNNs, the input layer and the following hidden layers learn a low-dimensional and high-level representation, then the rest of the hidden layers and the output layer work as the predictor using the representation. In communication systems, the encoder compresses the data as the signal for transmission, then the decoder reconstructs the original data using the signal. However, the aim of the decoder is the input of communication systems, while the predictor aims at the label rather than the input of DNNs. \citet{shwartz2017opening} presents that the data flow in DNNs forms a Markov chain, namely,
\begin{equation}
    I(X;Y)\geq I(T_1;Y) \geq \dots \geq I(T_k;Y) \geq I(Y';Y) 
\end{equation}
where $T_k$ represents the representation of the $k$-th hidden layer, and $\hat{Y}$ is the prediction. \citet{shwartz2017opening} and \citet{yu2020understanding} experimentally validated Markov chains on DNNs. Therefore, the label can be regarded as the information source for DNNs, but it has been implicitly processed as the input data. DNNs continue to resolve the label from the input data like communication systems, and each layer performs like either the encoder or decoder. Since the representation of one layer consists of the outputs of all this layer's neurons, the layer is the training atom in MOLE and some layers can be treated as a module for training. Since each layer of DNNs holds InfoMax \cite{linsker1988self}, each module is trained by maximizing the mutual information between its output and the desired variable. Since the output of one module is the input of its following module, each module is sequentially trained from the input module to the output module.

Intuitively, all modules in DNNs act as decoders that should perform optimal resolution for the labels, namely $I(T_k, Y)$ maximization. However, all the low-level features with equivalent importance are used to resolve, such that the DNN would be struggling to differentiate among the truly similar samples. Similarly to human recognition, we rarely correctly distinguish twins by their entire outlook but by their facial features. According to \citet{tian2020makes}, training by only $I(T_k; Y)$ maximization leads to substantial noise, causing the representation learning before the predictor vulnerable to adversarial attacks. Therefore, the first several layers should perform as the encoder to extract the important information and embed the low-level features as the high-level representation, and the following layers perform as the encoder to resolve the high-level representation as the label. In addition, the layer or module assigned for encoder-like modules should be trained by the training objective, $I(T_k; X)$ maximization, while the training objective for decoder-like modules is $I(T_k; Y)$ maximization. The pipeline of MOLE is shown in Figure \ref{pipline}. In principle, MI measures the similarity between two variables' space, so the trained model via MOLE would perform better on unobserved samples after $I(Y'; Y)$ is maximized on the output module. However, the optimization by gradient descent techniques approximates the maximum $I(Y'; Y)$ in the batch setting. Because this step is based on the approximation, there would exist groups of outputs with the almost same similarity to the labels if the output module is still trained by empirical $I(Y'; Y)$ maximization and only several groups have relatively good performance. From the geometrical interpretation, the model performs well enough if and only if the output's hyperplane most coverages to the label's hyperplane \cite{yu2019understanding}. To ensure that DNNs always have good performance, the conventional loss function replaces MI on the output module in the current state, and the conventional loss function is equivalent to $I(Y'; Y)$ maximization.
%\textcolor{green}{there would exist groups of outputs with the almost same similarity to the label, namely the extremely small difference of MI. So the accuracy of the model would be good enough if and only if the output's hyperplane (almost) has the largest projection on the label's hyperplane \cite{yu2019understanding}. To ensure that DNNs always have good accuracy, the conventional loss function replaces MI on the output module in the current state.}
As Figure \ref{pipline} shows, it is clear that using MOLE makes DNNs locally learn. MOLE allows DNNs to learn from unlabeled data for general high-level representation in the encoder-like module, and the few-shot empirical representations with labels would cluster through the decoder-like module. Therefore, MOLE remains biologically plausible. Furthermore, in the case of a DNN with a fixed number of layers, the allocation of a greater number of layers to the encode-like module and a smaller number of layers to the decode-like module enhances the impact of $I(T_k; X)$ while diminishing the influence of $I(T_k; Y)$, and vice versa. MOLE implicitly forms IB by module assignment such that DNNs trained by MOLE would be more robust to some extent.

%information bottleneck\cite{tishby2000information}

%\cite{tishby2015deep}

%\cite{shwartz2017opening}

%\cite{yu2020understanding}
\section{Experiment}

\begin{table*}[ht]
\centering
\caption{Classification accuracy results on \textit{Adult}, \textit{MNIST}, \textit{CIFAR-10}, \textit{Mutagenicity}, \textit{Cora} with different estimators}
\vspace*{3mm}
\small
\begin{tabular}{@{}cccccc@{}}

\toprule
Dataset                       & Estimator     & Train Acc. & Diff. of Train Acc. to BP & Test Acc. & Diff of Test Acc. to BP \\ \midrule
\multirow{3}{*}{\textit{Adult}}        & BP            & 91.08\%        & /                            & 83.94\%       & /                          \\
                              & Matrix-based  & 84.74\%        & -6.34\%                      & 84.79\%       & 0.85\%                     \\
                              & MINE          & 83.58\%        & -7.5\%                       & 83.69\%       & -0.25\%                      \\ \midrule
\multirow{4}{*}{\textit{MNIST}}        & BP            & 99.94\%        & /                            & 99.28\%       & /                          \\
                              & Matrix-based  & 96.65\%        & -3.29\%                      & 94.71\%       & -4.57\%                    \\
                              & MINE          & 95.05\%        & -4.89\%                      & 94.49\%       & -4.79\%                    \\
                              & DIM+MINE      & 98.60\%        & -1.34\%                      & 96.85\%       & -2.43\%                    \\ \midrule
%\multirow{4}{*}{CIFAR-10}     & BP            & 94.39\%        &       /                       & 69.59\%       &           /                 \\
                   %           & Matrix-based  & 53.49\%        &   -40.90\%                           & 43.26\%       &    -26.33\%                         \\
                    %          & MINE          &  80.35\%              &        -14.04\%                      &    50.68\%            &   -18.91\%                         \\
                     %         & DIM+MINE      &   80.188\%             &                              &     53.59\%          &                            \\ \midrule
\multirow{4}{*}{\textit{Mutagenicity}} & BP            & 76.82\%        & /                            & 77.63\%       & /                          \\
                              & Matrix-based &   67.90\%             &             -8.92\%                &   66.55\%            &  -11.08\%                          \\
                              & MINE          & 70.63\%        & -6.19\%                      & 69.55\%       & -8.08\%                    \\
                              & GMI+MINE      & 75.56\%        & -1.26\%                      & 76.01\%       & -1.62\%                    \\ \midrule
\multirow{2}{*}{\textit{Cora}}         & BP            & 99.29\%        & /                            & 70.60\%       & /                          \\
                              & GMI+MINE           & 82.86\%        & -16.43\%                     & 68.40\%       & -2.20\%                    \\ \bottomrule
\end{tabular}
\label{experiment_summary}
\end{table*}

This section experimentally shows that MOLE is universally applicable. Since MOLE is proposed by analogizing communication systems, and each module works as a data processor, MOLE generally works as long as it can process all types of data. Considering the most commonly-used data types in DNNs, i.e., vector, grid, and graph, we utilize these datasets: \textit{Adult} (vectors), \textit{MNIST} (grids), \textit{Cora} (a graph) and \textit{Mutagenicity} (graphs). We implement Multi-layer Perceptrons \citep[MLP,][]{murtagh1991multilayer} on \textit{Adult}, Convolutional Neural Networks \citep[CNN,][]{lecun1998gradient} on \textit{MNIST}, Message-Passing Graph Neural Networks  \citep[MPGNN,][]{gilmer2017neural} on \textit{Mutagenicity}, and Graph Convolutional Networks  \citep[GCN,][]{kipf2016semi} on \textit{Cora}. 

%For \textit{Adult}, the Multi-layer Perceptrons contain 3 fully-connected layers. The network for \textit{MNIST} has 2 convolutional layers and 2 fully connected layers. We implement the Message-Passing Neural Network \cite{gilmer2017neural} on \textit{Mutagenicity}: one  fully-connected layer for embedding, two message-passing layers, and one fully-connected layer for prediction. Graph Convolutional Networks\cite{kipf2016semi} is used for \textit{Cora}: 2 graph convolutional layers, and 1 fully-connected layer.

In the training, we directly use the high-dimension MI estimators to achieve InfoMax instead of the well-performed contrast learning method or other alternatives. Although such estimators are highly computational and even cannot capture implicit information well, 
%for example, the spatial locality in the grid-like data, 
it definitively conforms to the underlying mechanism of communication systems, and straightforwardly shows the quality of the measure mainly affects the final outcomes of MOLE. We implement two main types of estimators, parametric and non-parametric. The non-parametric is the matrix-based estimator \cite{yu2021measuring}, which can consistently measure between the vector-type data. We further use the multivariate extension \cite{yu2020understanding} to extend the matrix-based estimator on the grid- and graph-type data. However, the grid extension estimator does not consider the spatial locality, while the graph extension can partially capture the topological connectivity by involving the adjacent vector. We use parametric estimators including MINE, DIM, and GMI. In detail, MINE performs on the vector-type representation well, as the lower bound is very tight. DIM can capture more information about spatial locality by involving local MI. DGI captures the topological connectivity by learning the topology-aware MI. 

%We experimentally show the viability compared with BP, so we use naive neural networks. 
The assignment of the training objective and the choice of DNNs are shown in Appendix \ref{experiment setup}. The results are tabulated in Table \ref{experiment_summary}. For \textit{Adult}, we run three training: BP, MOLE with Matrix-based estimator, and with MINE. Their accuracies on the unobserved data are close, 83.94\%, 84.79\%, and 83.69\% respectively. MOLE with Matrix-based estimator has the highest test accuracy, which is even higher than its train accuracy. In \textit{MNIST}, the performance of the extended Matrix-based estimator and MINE on MOLE are clearly lower than BP, and the test accuracy differences reach over 4\%. The object in \textit{MNIST} is to recognize a grayscale handwritten number on a black background, which means the images have relatively weak spatial locality. If the object and background become more complex with stronger spatial locality, the accuracy difference would enhance. Using DIM to train the encode-like modules narrows the difference. In \textit{Mutagenicity}, 
%we also implement MOLE with only MINE (MINE only measures the information of node features without the topological connections). 
the test accuracies on MOLE with Matrix-based estimator, MINE, and GMI+MINE are 66.66\%, 69.55\%, and 76.01\%, while BP's is 77.63\%. With the topological connectivity considered by GMI, the model outperforms other models with MOLE, approaching BP. \textit{Cora} is used for the semi-supervised node-level task in GNN, and MOLE with GMI+MINE has close test accuracy to BP. 
Based on the experiment results, 
%we can conclude that the more information the estimator captures, the closer MLP performs to BP. In other words, 
we can conclude the quality of the MI estimator decisively affects the performance of neural networks. Compared with the train and test accuracies on MOLE, we observe that both are very close for most types of neural networks even though the test accuracy is higher. Therefore, we believe that MOLE would improve the generalization of neural networks. Although GCN on \textit{Cora} with MOLE has a large difference between the train and test accuracies, BP even makes this model over-fitting. Therefore, this large difference is reasonable.

After training, we visualize the distribution of samples from raw data to each layer's output in the 2-dimensional plane by t-SNE \cite{van2008visualizing} embedding as shown in Appendix \ref{visualization}. Ideally, we hope the representation would start to cluster by classes in the encoder-like layer as Figure \ref{Cora} shows. In most conditions, the output of the encoder-like layer would have a similar or symmetrically similar distribution to the raw data, and the output of the decoder-like layer would cluster by class. For instance, \textit{MNIST} significantly fulfills this condition as shown in Figure \ref{MNIST}. So the effect of $I(T_k; Y)$ maximization is to cluster the representations. Therefore, the maximal distance across classes would be an alternative measure to train the decoder-like layer with less computation. After the last layer trained by the conventional loss function, two phenomenons would happen: the samples cluster by class further with less overlapping in the distribution as shown in Figure \ref{Mutag} or distance across cluster enhanced as shown in Figure \ref{MNIST}, and the samples and classes become nonlinearly separable in Figure \ref{Adult}. Specifically, the last layer trained by the conventional loss function would boost the cluster in decoder-like layers.

%\cite{yu2020understanding}

%\cite{yu2019understanding}
\section{Discussion}
Recently, \citet{hinton2022forward} clarifies the drawbacks of BP and proposes the alternative Forward-Forward (FF) algorithms inspired by contrastive Learning \cite{gutmann2010noise} and Boltzmann machines \cite{hinton1986learning}. In FF, the label encoding is added to the input data. The input with correct encoding forms positive samples, while the input with incorrect encoding forms negative samples. FF uses the measure of goodness as the training objective. When the positive sample is forwarded, the parameters are updated by maximizing the goodness, and vice versa. The operation of two forward passes equivalently is to resolve the label from the input, namely hidden modules are trained by maximizing $I(T_k; Y)$. \citet{paliotta2023graph} extends FF on the graph domain, named GFF, but GFF only adapts the graph-level task. \citet{duan2021modularizing} trains the hidden module by maximizing the inter-class dissimilarity of hidden outputs, and the class of the hidden output is decided by the label of the input. In other words, after training, the hidden output should cluster by the label, which has the same effect of $I(T_k; Y)$ maximization. Similarly, Associated Learning \cite{wu2022associated} adds an autoencoder for the label, and the latent variable has the same dimension as the hidden output. The training mainly depends on the minimization of the difference between the hidden output and the latent variable. However, these approaches would still face the problem of over-reliance on labels, and adding the equivalently $I(T_k, X)$-maximization operation would bring improvement.

\citet{lowe2019putting} propose the Greedy InfoMax (GIM) approach for unsupervised representation learning, equivalently encoder-like modules trained by $I(T_k, X)$ maximization. \citet{ma2020hsic} replaces MI as the optimized measure by HSIC Bottleneck trading-off between the information the hidden representation needs for predicting the output and the information the hidden representation retains about the input. Module assignment in MOLE between encoder- and decoder-like training is such an explicit trade-off. Overall, these backpropagation-free approaches fulfill our proposed MOLE. In addition, MOLE still remains largely unexplored, and it would experience long-time research and exploration to outperform BP. For example, piecewise-linear functions, such as ReLU and LeakyReLU are used to prevent the gradient vanishing and exploding, but these would achieve limited non-linear transformation. The output from the linear transformation is equivalent to the signal in communication systems. This output should be non-linearly transformed as a whole. So using the kernel method \cite{zhang2017stacked, duan2021modularizing} would be a more suitable non-linear transformation for MOLE. In the semi-supervised setting, the imbalanced measure between the output and the label on the decoder-like module would be helpful as well. Furthermore, MOLE would achieve parallel training in the encoder-like modules by designing the parallel architecture in DNNs. Each parallel architecture is trained by maximizing $I(T_k;X)$ and their outputs are concatenated as the input of the following decoder-like module.

%\begin{itemize}
    %\item satisfies Biologically plausible: 1. local optimization; 2. semi-supervised classification children can learn to recognize a new category based on a handful of samples.
    %\item (possible) the existing technique also satisfies this framework, distinguishinging the positive and negative in FF
%\end{itemize}
%FF\cite{hinton2022forward}: contrast learning make the input coverage to the label like resolving the label from the input (only I(T; Y) maximization)

%GFF\cite{paliotta2023graph} extend FF to graph. But only workable on graph-level task edge-level?

%\cite{lowe2019putting} InfoNCE performs as the lower bound of MI. The paper performs the encoder-like optimization. After following the predictor, from the high-level representation, the acc is good

%\cite{ren2022scaling}

%kernel\cite{duan2021modularizing} redefine DNNs,  the label requirement of deep learning

%Focusing on the two-module case, we prove that the training of input and output modules can be decoupled by leveraging pairwise kernel evaluations on training examples from distinct classes.

%the maximal distance between distinct classes for the hidden layers, which has the same effect of I(Y;Z) maximization

%Associated Learning\cite{wu2022associated} mapping the label to the latent variable with the same dimension as the hidden output. minimize the dissimilarity between the hidden output and the latent variable (I(Y;T) maximization) 

%HSIC Bottleneck\cite{ma2020hsic} use HSIC bottleneck as the measure in the hidden layer. the trade-off, although local optimization, all layers are trained together 
\section{Conclusion}

We introduce Modular Learning Framework, which trains DNNs with gradient-isolated across modules. Although currently, MOLE can not ultimately outperform BP, it deserves further research because of its apparent advantages (biological plausibility, universal applicability, etc). These advantages would help the interpretability of DNNs, for example, it would make the outcome-based interpretation more reliable and reasonable.

During the investigation of MOLE, we also find some questions worth further exploration:
\begin{itemize}
    \item Can we find the optimal module assignment between the encoder- and decoder-like components?
    \item In the out-of-experiment setting, the object and background in the grid-type data are much more complex, so the data mainly depends on the spatial locality. Can we design an estimator to better capture the implicit spatial locality?
    \item The MI estimator for the high-dimensional data is usually time-consumed and not accurate enough, particularly in the encoder-like module. Can we design an alternative technique to achieve InfoMax in the decode-like module? For example, certain self-supervised representation learning methods \citep[SSRL,][]{ericsson2022self} would work. If it works, this approach could match the accuracy of backpropagation (BP) and significantly expedite the learning process. Furthermore, for more complex data such as audio and video, can we also achieve InfoMax by SSRL? 
    \item After the application of MOLE, each module becomes independent. Is it possible to prove how the parameters of each module span the hyperplane of its output?
    \item If the span of the hyperplane succeeds, can we design an optimization algorithm that employs geometric projection techniques to obtain an exact optimal solution?
\end{itemize}
These questions present exciting avenues for further research and could potentially enhance our understanding and utilization of MOLE.

% Acknowledgements should only appear in the accepted version.
%\section*{Acknowledgements}

% In the unusual situation where you want a paper to appear in the
% references without citing it in the main text, use \nocite

\bibliography{example_paper}
\bibliographystyle{icml2023}

%%%%%%%%%%%%%%%%%%%%%%%%%%%%%%%%%%%%%%%%%%%%%%%%%%%%%%%%%%%%%%%%%%%%%%%%%%%%%%%
%%%%%%%%%%%%%%%%%%%%%%%%%%%%%%%%%%%%%%%%%%%%%%%%%%%%%%%%%%%%%%%%%%%%%%%%%%%%%%%
% APPENDIX
%%%%%%%%%%%%%%%%%%%%%%%%%%%%%%%%%%%%%%%%%%%%%%%%%%%%%%%%%%%%%%%%%%%%%%%%%%%%%%%
%%%%%%%%%%%%%%%%%%%%%%%%%%%%%%%%%%%%%%%%%%%%%%%%%%%%%%%%%%%%%%%%%%%%%%%%%%%%%%%
\newpage
\appendix
\onecolumn

\section{Appendix}\label{appendix}
\subsection{Datasets}
\label{appendix_datasets}
In the experiment, we use the public and accessible dataset, and \textit{MNIST}, and \textit{Cora} are well-known and common-used datasets for experiments. Here, we show the details of \textit{Adult} and \textit{Mutagenicity} datasets.

\textit{Adult}\footnote{https://archive.ics.uci.edu/ml/datasets/Adult} has 48842 samples, of which 3620 samples with \textit{null} values exist. \textit{Adult} is used for
salary classification, so salary is used as the label. salary is a binary variable: \texttt{>50K} and \texttt{<=50K}. Table \ref{tab:adult} shows 14 features and their data types in \texttt{Adult}. After preprocessing (filter out the samples with \textit{null}, encode the category data by one-hot encoding), there are 104 numerical features.

\textit{Mutagenicity}\footnote{https://chrsmrrs.github.io/datasets/docs/datasets/} is a graph dataset, and each sample consists of a graph and a binary label. The graph describes the structure of molecules, and the binary label indicates whether a molecule can induce genetic mutations. \textit{Mutagenicity} is used for graph classification based on the features of nodes and structures of the graphs, and it contains 4,337 graphs that all consist of 14 basic elements.

\begin{table}[htbp]
\centering
\caption{The attributes of \textit{Adult} }
\onecolumn
\begin{tabular}{cc}
\hline
\textbf{Attribute}      & \textbf{Data Type} \\ \hline
\texttt{age}            & integer   \\ 
\texttt{workclass}      & category  \\ 
\texttt{fnlwgt}         & integer   \\ 
\texttt{education}      & category  \\ 
\texttt{education-num}  & category  \\ 
\texttt{marital-status} & category  \\ 
\texttt{occupation}     & category  \\ 
\texttt{relationship}   & category  \\ 
\texttt{race}           & category  \\ 
\texttt{sex}            & category  \\ 
\texttt{capital-gain}   & integer   \\ 
\texttt{capital-loss}   & integer   \\ 
\texttt{hours-per-week} & integer   \\ 
\texttt{native-country} & category  \\ 
 \hline
\end{tabular}\label{tab:adult}
\end{table}

\subsection{Experiment Setup}
We experimentally show the viability compared with BP, so we use naive neural networks, stacking 3 to 4 layers.
\label{experiment setup}
\paragraph{Multilayer Perceptrons (MLP) on \textit{Adult} Dataset} A three-layer MLP backbone is structured, and it is implemented on \textit{Adult} dataset. The first layer has 64 perceptions for 104-dimensional input, and the 64-dimensional output is fed to the second layer with 16 perceptrons. All the perceptrons are followed by ReLU activations for non-linearity. The last layer consists of two perceptrons and uses SoftMax activation to predict the probability of each class. The first layer is trained by $I(T_k; X)$ maximization, the second layer is trained by $I(T_k, Y)$, and the last layer is trained by minimizing the cross entropy.

\paragraph{Convolutional Neural Networks (CNN) on \textit{MNIST} Dataset} A four-layer Convolutional Neural Network is initialized: two convolutional layers and two fully-connected layers. It is clear that the convolutional layer is to extract the high-level representation from images, and both are trained by $I(T_k; X)$ maximization. The first fully-connected layer is trained by $I(T_k;Y)$ maximization, and the second is trained by minimizing the cross entropy. When we use DIM+MINE as the MI estimator, we use DIM to estimate the MI between grids and grids on the first convolutional layer, use MINE to estimate the MI between grids and vectors on the second convolutional layer, and use MINE to estimate the MI between vectors and vectors on the first fully-connected layer.

\paragraph{Message Passing Graph Neural Networks (MPGNN) on \textit{Mutagenicity} Dataset}
A four-layer Message Passing Graph Neural Network is used. The first layer performs the node embedding: a fully-connected layer on the feature matrix, and it is trained by $I(T_k; X)$ maximization. The following layer is a message-passing (MP) layer trained by $I(T_k; X)$ maximization as well. And the next message-passing (MP) layer is trained by $I(T_k; Y)$. And the final layer is a fully-connected with the cross entropy minimized. When we only use MINE as the MI estimator, MINE is to estimate the MI between vectors and vectors. In other words, we only consider the node features. When we use GMI+MINE as the MI estimator, we use MINE to estimate the MI between vectors and vectors on the embedding layer, use GMI to estimate the MI between graphs and graphs on the first message-passing layer, use MINE to estimate the MI between vectors and vectors on the second message-passing layer.

\paragraph{Graph Convolutional Networks (GCN) on \textit{Cora} Dataset}
A three-layer Graph Convolutional Network has been implemented: two graph convolutional layers and one fully-connected layer. The first layer is trained by $I(T_k; X)$, the second is trained by $I(T_k;Y)$ where the measure only considers MI between the representation with the known label and their label, and the last is trained by minimizing the cross entropy. When we use GMI+MINE as the MI estimator, we use GMI to estimate the MI between two graphs on the first graph convolutional layer and use MINE to estimate the MI between vectors and vectors on the second graph convolutional layer.

\subsection{Visualization}\label{visualization}
\begin{figure}[!]
\centering
\subfigure[Visualization of raw features' t-SNE embeddings]{
\includegraphics[width=8cm]{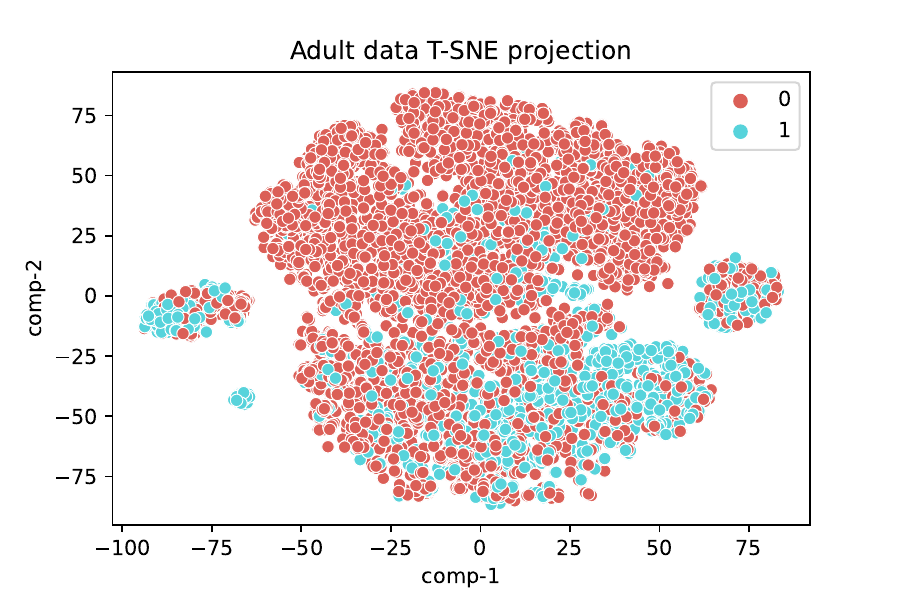}
%\caption{fig1}
}
\quad
\subfigure[Visualization of the first-layer outputs' t-SNE embedding]{
\includegraphics[width=8cm]{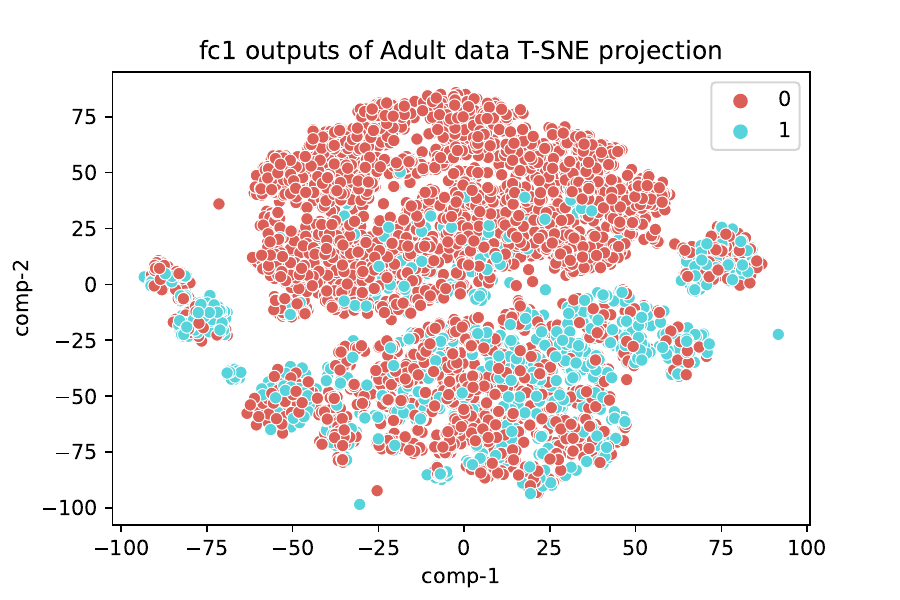}
}
\quad
\subfigure[Visualization of the second-layer outputs' t-SNE embedding]{
\includegraphics[width=8cm]{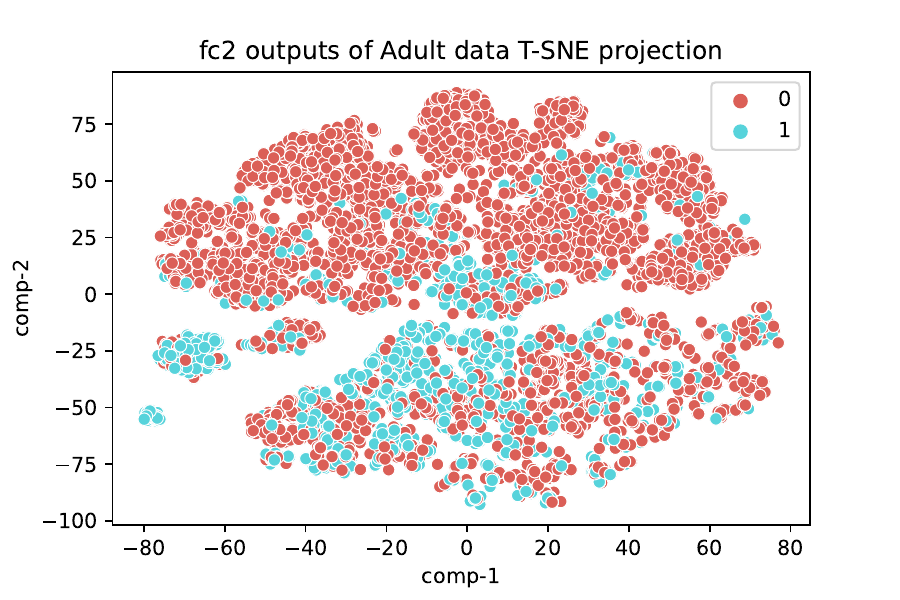}
}
\quad
\subfigure[Visualization of the last-layer outputs' t-SNE embedding]{
\includegraphics[width=8cm]{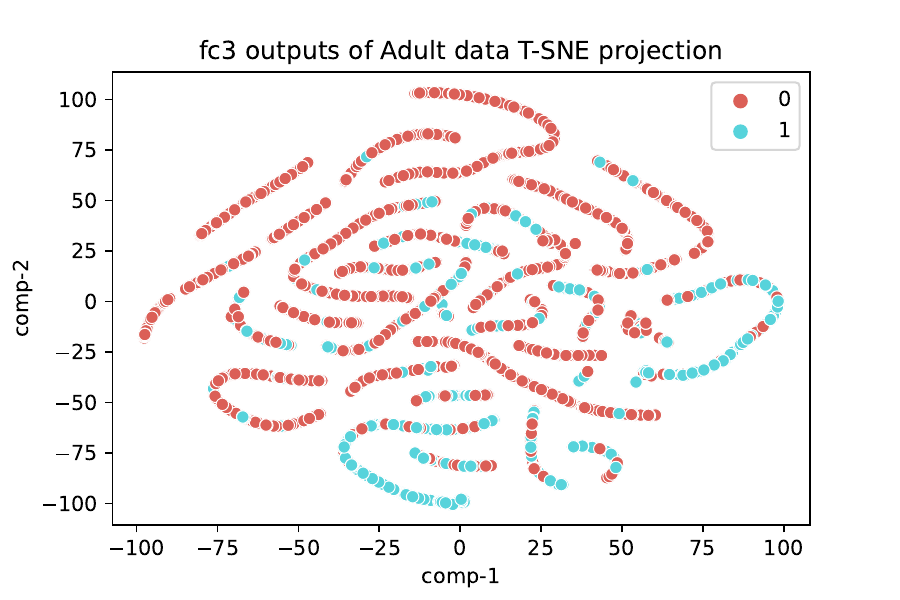}
}
\vspace*{3mm}
\caption{Visualization of t-SNE embeddings from raw features of \textit{Adult} dataset to each output of MLP}
\vspace*{3mm}
\label{Adult}
\end{figure}

\begin{figure}[!]
\centering
\subfigure[Visualization of raw features' t-SNE embeddings]{
\includegraphics[width=8cm]{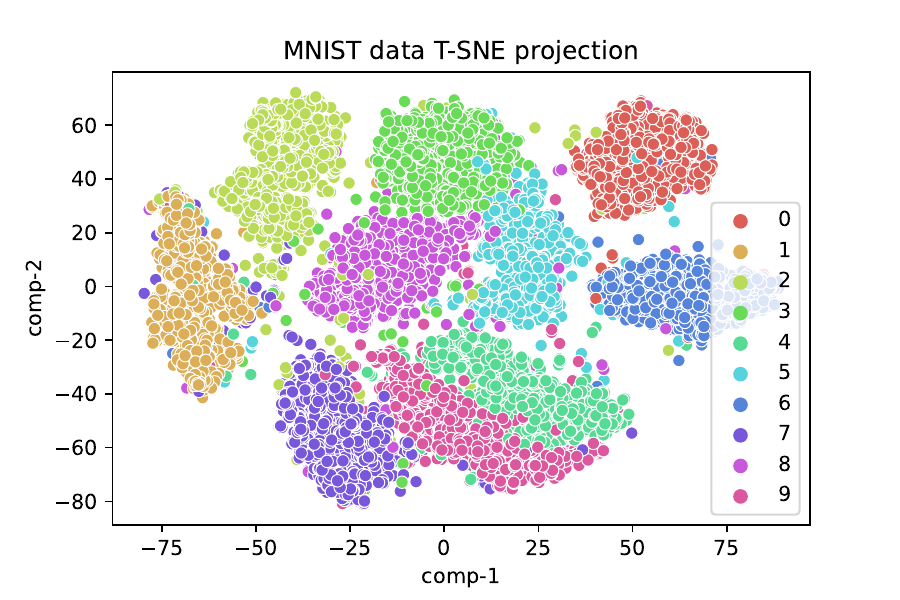}
%\caption{fig1}
}
\quad
\subfigure[Visualization of the first-layer outputs' t-SNE embedding]{
\includegraphics[width=8cm]{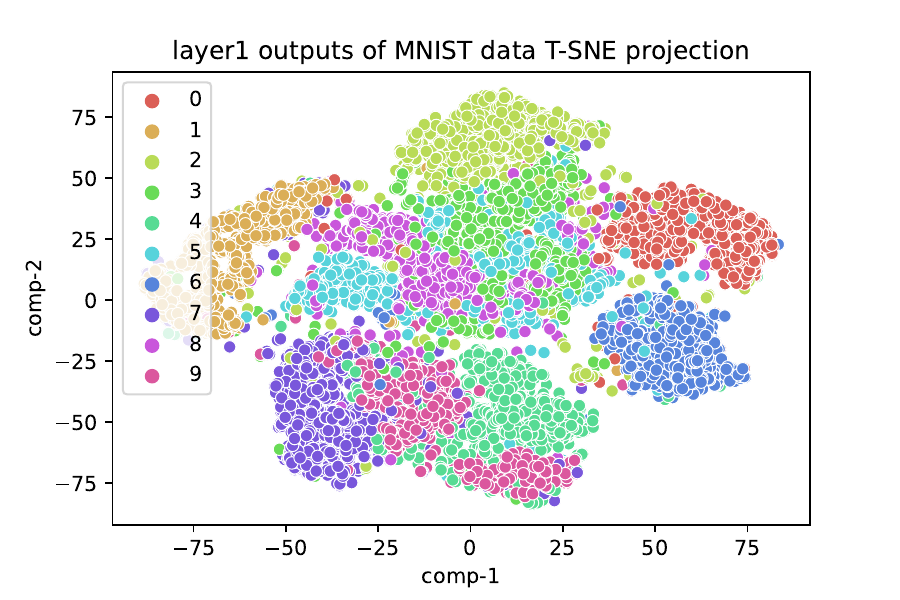}
}
\quad
\subfigure[Visualization of the second-layer outputs' t-SNE embedding]{
\includegraphics[width=8cm]{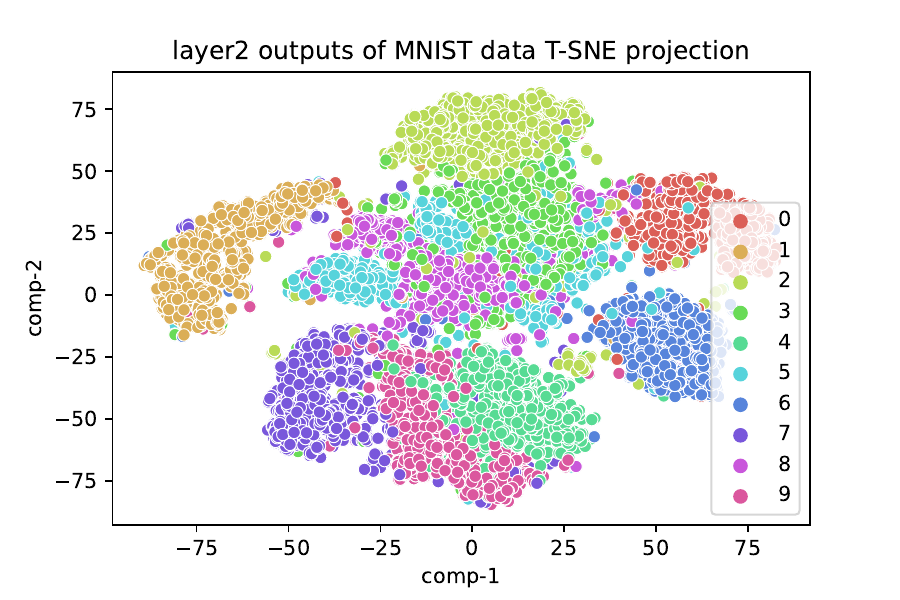}
}
\quad
\subfigure[Visualization of the third-layer outputs' t-SNE embedding]{
\includegraphics[width=8cm]{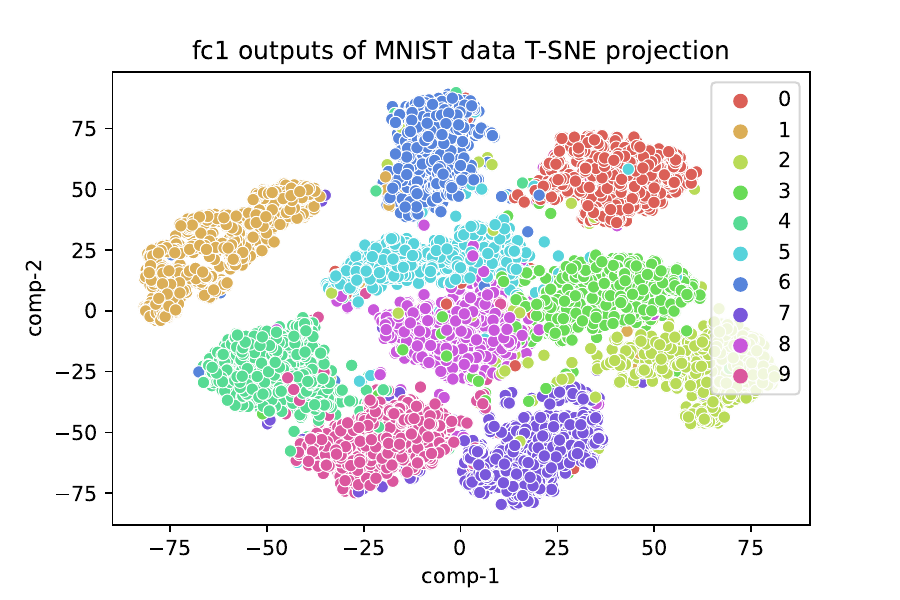}
}
\quad
\subfigure[Visualization of the last-layer outputs' t-SNE embedding]{
\includegraphics[width=8cm]{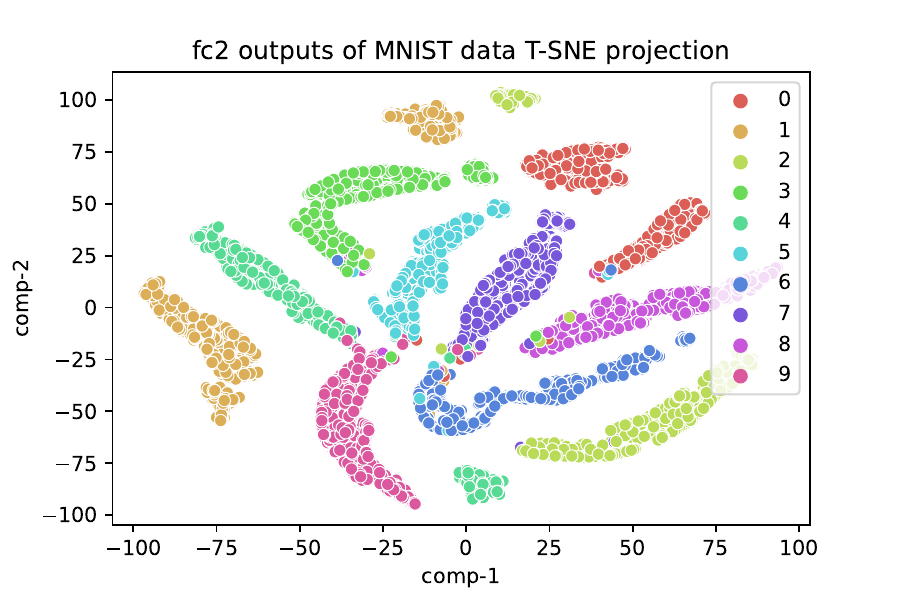}
}
\vspace*{3mm}
\caption{Visualization of t-SNE embeddings from raw features of \textit{MNIST} dataset to each output of CNN}
\vspace*{3mm}
\label{MNIST}
\end{figure}

\begin{figure}[!]
\centering
\subfigure[Visualization of the second MP-layer outputs' t-SNE embedding]{
\includegraphics[width=8cm]{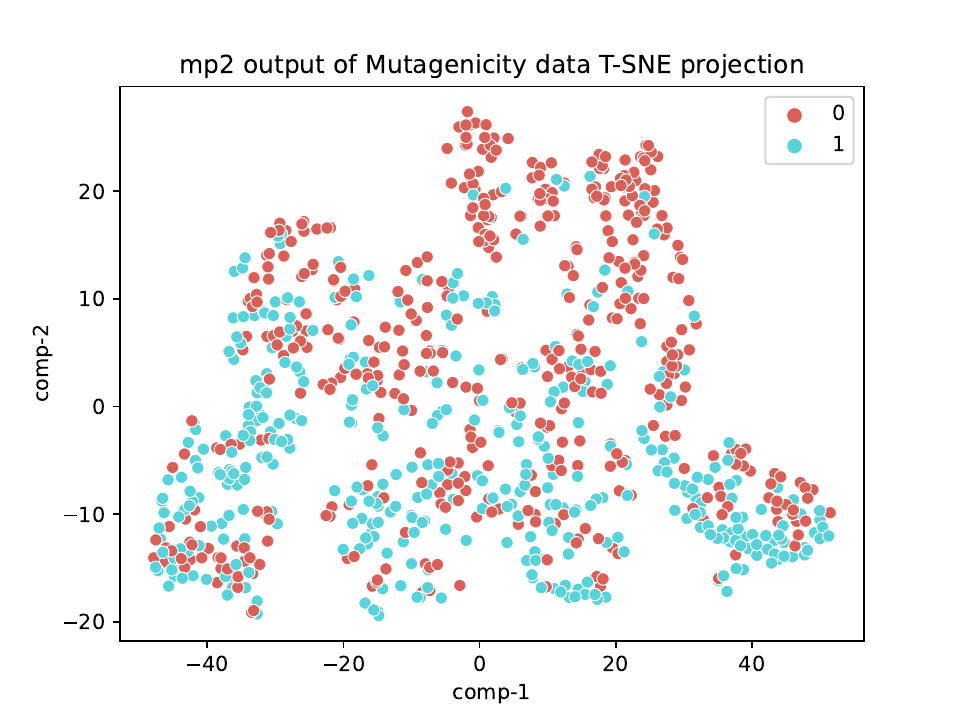}
%\caption{fig1}
}
\quad
\subfigure[Visualization of the last-layer outputs' t-SNE embedding]{
\includegraphics[width=8cm]{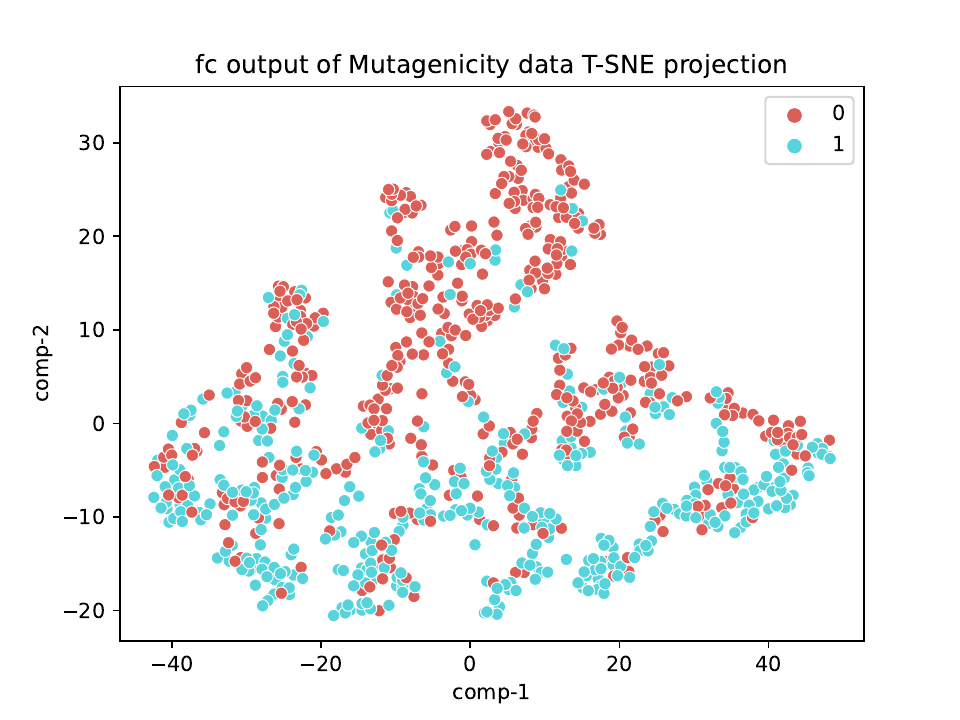}
}
\vspace*{3mm}
\caption{Visualization of t-SNE embeddings from raw features of \textit{Mutagenicity} dataset to each output of MPGNN}
\vspace*{3mm}
\label{Mutag}
\end{figure}

\begin{figure}[!]
\centering
\subfigure[Visualization of raw features' t-SNE embeddings]{
\includegraphics[width=8cm]{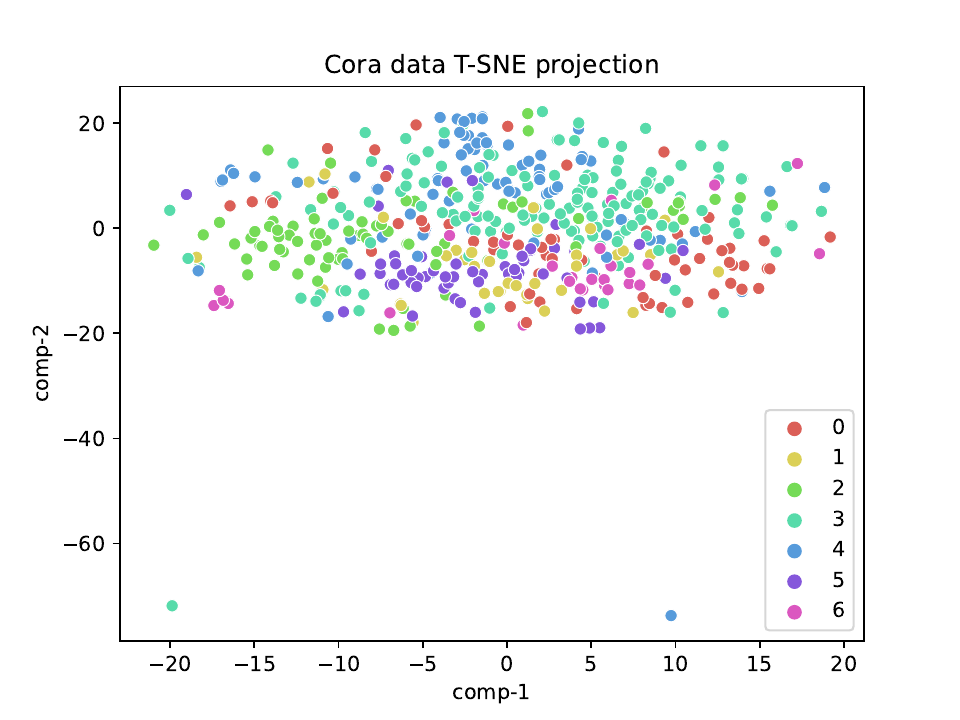}
%\caption{fig1}
}
\quad
\subfigure[Visualization of the first-layer outputs' t-SNE embedding]{
\includegraphics[width=8cm]{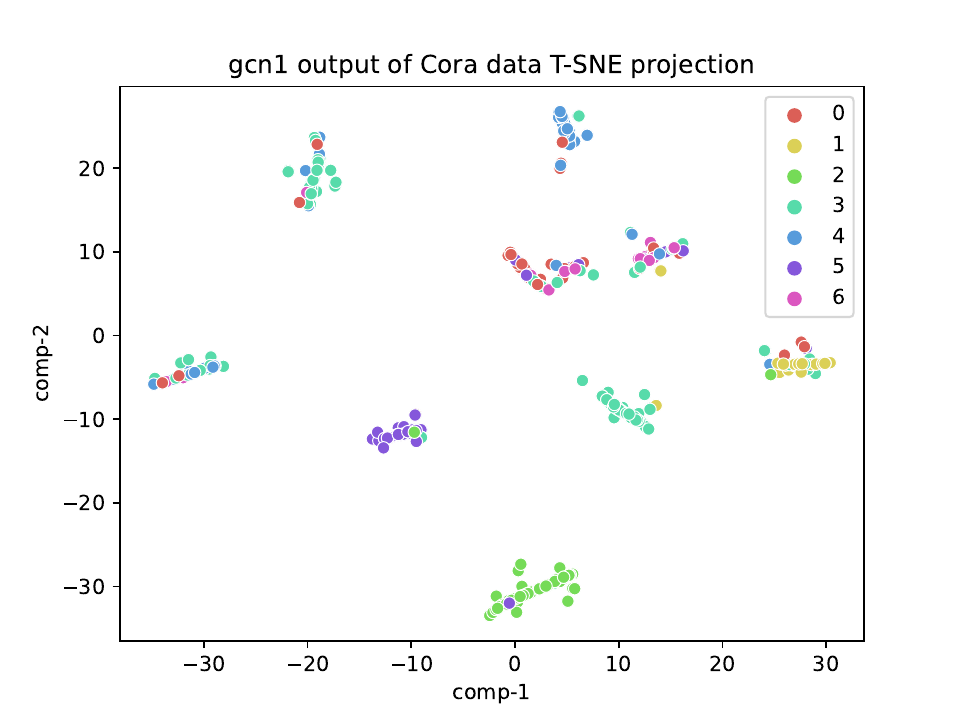}
}
\quad
\subfigure[Visualization of the second-layer outputs' t-SNE embedding]{
\includegraphics[width=8cm]{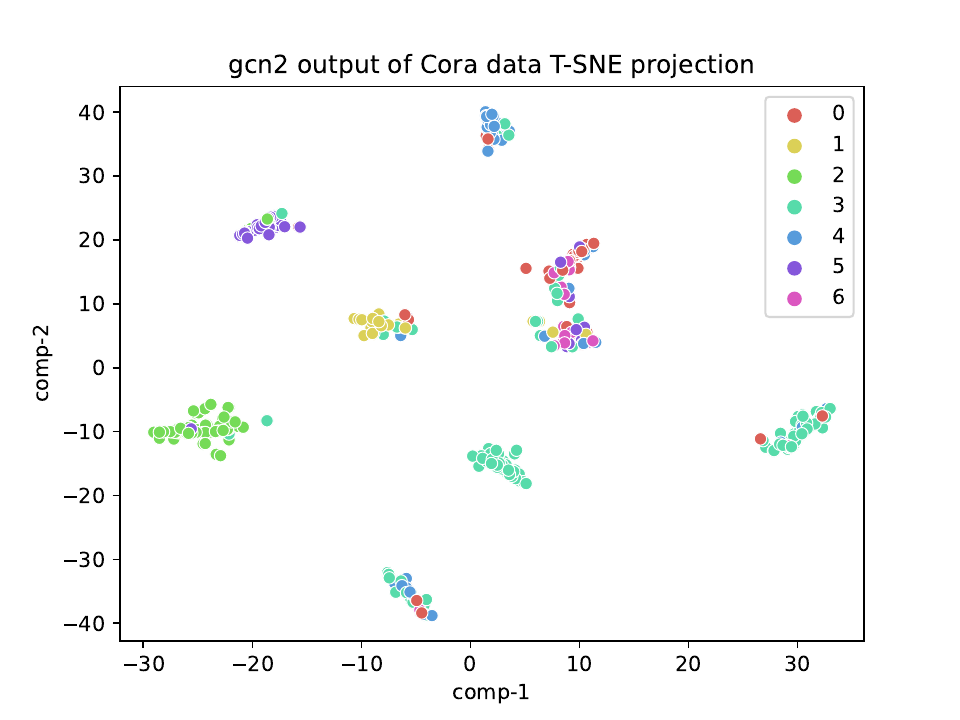}
}
\quad
\subfigure[Visualization of the third-layer outputs' t-SNE embedding]{
\includegraphics[width=8cm]{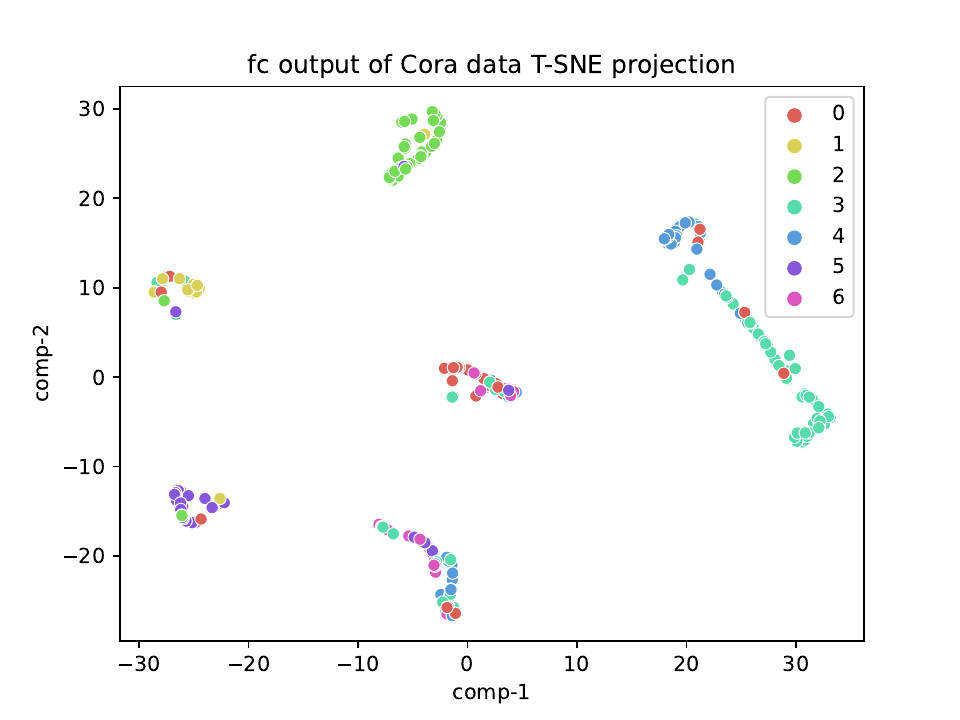}
}
\vspace*{3mm}
\caption{Visualization of t-SNE embeddings from raw features of \textit{Cora} dataset to each output of GCN}
\vspace*{3mm}
\label{Cora}
\end{figure}

Here, we use t-SNE to embed the high-dimensional data as the 2-dimensional and visualize the embeddings in the 2-dimensional plane. Figure \ref{Adult} shows the distribution of the raw samples from \textit{Adult} and the corresponding outputs from each layer, and Figure \ref{MNIST} and Figure \ref{Cora} respectively show \textit{MNIST}'s and \textit{Cora}'s.
Figure \ref{Mutag} only the visualization of the last two layers trained by maximizing $I(T_k; Y)$ and minimizing the cross entropy respectively, because the raw samples and the samples' outputs from the first two layers have different dimensions of features.

\end{document}